\documentclass[sigconf]{acmart}

\acmConference[none]{none}{none}{none}

\usepackage{algorithm}
\usepackage[noend]{algpseudocode}
\usepackage{graphicx}
\usepackage{textcomp}

\usepackage{subfigure} 
\usepackage{soul}
\usepackage{booktabs}
\usepackage{diagbox}
\usepackage{threeparttable}
\usepackage{array}
\usepackage{balance}
\usepackage{hyperref}
\usepackage{url}
\usepackage{color}
\usepackage{xcolor}
\usepackage{enumitem}
\usepackage{tcolorbox}

\usepackage[justification=centering]{caption}
\usepackage{float}
\usepackage{hhline}
\usepackage{multirow}
\usepackage{xspace}

\def \projectName {CNNSplitter\xspace}

\setcopyright{none}
\renewcommand\footnotetextcopyrightpermission[1]{} 

\begin{document}
\title{Patching Weak  Convolutional Neural Network Models through Modularization and Composition}

\author{Binhang Qi*}
\affiliation{
  \institution{SKLSDE Lab, Beihang University\country{China}}
}
\email{binhangqi@buaa.edu.cn}

\author{Hailong Sun*\textsuperscript{\textdagger}}
\affiliation{
  \institution{SKLSDE Lab, Beihang University\country{China}}
}
\email{sunhl@buaa.edu.cn}

\author{Xiang Gao*\textsuperscript{\textdagger}}
\affiliation{
  \institution{SKLSDE Lab, Beihang University\country{China}}
}
\email{xiang_gao@buaa.edu.cn}

\author{Hongyu Zhang}
\affiliation{
  \institution{The University of Newcastle\country{Australia}}
}
\email{hongyu.zhang@newcastle.edu.au}

\begin{abstract}
Despite great success in many applications, deep neural networks are not always robust in practice. 
For instance, a convolutional neuron network (CNN) model for classification tasks often performs unsatisfactorily in classifying some particular classes of objects. In this work, we are concerned with patching the weak part of a CNN model instead of improving it through the costly retraining of the entire model. 
Inspired by the fundamental concepts of modularization and composition in software engineering, we propose a compressed modularization approach, \projectName, which decomposes a strong CNN model for $N$-class classification into $N$ smaller CNN modules. 
Each module is a sub-model containing a part of the convolution kernels of the strong model. 
To patch a weak CNN model that performs unsatisfactorily on a target class (TC), we compose the weak CNN model with the corresponding module obtained from a strong CNN model. 
The ability of the weak CNN model to recognize the TC can thus be improved through patching. 
Moreover, the ability to recognize non-TCs is also improved, as the samples misclassified as TC could be classified as non-TCs correctly.
Experimental results with two representative CNNs on three widely-used datasets show that the averaged improvement on the TC in terms of precision and recall are 12.54\% and 2.14\%, respectively. 
Moreover, patching improves the accuracy of non-TCs by 1.18\%. 
The results demonstrate that \projectName can patch a weak CNN model through modularization and composition, thus providing a new solution for developing robust CNN models.

\end{abstract}

\keywords{Modularization and Composition, DNN, CNN, Patching}

\maketitle

\renewcommand{\thefootnote}{\fnsymbol{footnote}}
\footnotetext[1]{Also with Beijing Advanced Innovation Center for Big Data and Brain Computing, Beihang University, Beijing 100191, China.}
\footnotetext[2]{Corresponding authors: Hailong Sun and Xiang Gao.}

\section{Introduction}
\label{sec:intro}
Modularization and composition are fundamental concepts in software engineering,
which facilitate software development, reuse, and maintenance by dividing an entire software system into a set of smaller modules.
Each module is capable of carrying out a certain task or separating a certain concern.  
For instance, when debugging a buggy program, testing and patching the module that contains the bug will be much easier than analysing the entire program.

Recently, convolutional neural networks (CNNs) have become one of the most effective deep learning models for processing a variety of tasks, such as image classification~\cite{alexnet}, object detection~\cite{girshick2014rich}, and semantic segmentation~\cite{long2015fully}. 
However, obtaining a strong CNN model with high accuracy is still very challenging, as it requires a large amount of quality data, careful data preprocessing, appropriate model structure, and proper model training and hyperparameter tuning strategies. 
A small problem %
in the model construction process
could result in a weak model (e.g., underfitting or overfitting model) with low accuracy. 
To improve the accuracy of weak models, %
the CNN models often need to be retrained with new data, model structure, training strategies, or hyperparameter values.
As the neural networks are getting deeper and the numbers of parameters and convolution operations are getting larger, the time and computational resources required for training the CNN models are rapidly growing~\cite{kirsch2018modular,li2016pruning}.

At a conceptual level, a CNN model is analogous to a program, and a mis-prediction of a CNN model is analogous to a program failure~\cite{ma2018mode, pei2017deepxplore}. %
Inspired by the application of modularization and composition in program debugging, it is natural to ask %
\textit{can the concepts of modularization and composition be applied to CNN models and facilitate the improvement of weak CNN models?} 
Through modularization and composition, the weak modules in a weak CNN model can be identified and patched separately; thus, the weak model can be improved without costly retraining the entire model. %

However, decomposing CNN models into modules is challenging:
(1) Different from traditional programs where each statement is readable, neural network models are composed of a set of neurons connected by uninterpretable parameter matrix.
Without fully understanding the effect and function of each parameter, decomposing models into different modules is hard.
(2) The connections between neurons in a neural network are complex and dense, hence it is hard to identify the relationship between neurons and prediction tasks (i.e., classify neurons into different modules).
Recently, researchers have studied techniques that decompose a fully connected neuron networks (FCNNs) model for $N$-class classification into $N$ modules, one for each class in the original model~\cite{fse2020modularity}. 
They achieved model decomposition by \textit{uncompressed modularization}, which removes a part of the weight of the neurons from a trained FCNN model.
However,  
due to the weight sharing~\cite{cnn2018overview,lenet} in CNNs, the relationship between weights and neurons in CNNs is many-to-many rather than many-to-one as in FCNNs. 
Removing the weight for one neuron in CNNs will influence all other neurons as well. 
As a result, this approach~\cite{fse2020modularity} cannot be applied to CNN models.
Although the follow-up work~\cite{nnmodularity2022icse} can be applied to decompose CNN models, it is still an uncompressed modularization approach.
The uncompressed modularization~\cite{fse2020modularity,nnmodularity2022icse} sets individual neurons or weights to 0, which does not change a module's structure and results in a module with the same number of weights as the original model
(to be discussed in Section \ref{subsec:struectued}).

To address the above challenges, in this paper, we propose a \textit{compressed modularization} approach, named \projectName, to decompose a CNN model into separate modules. 
Inspired by the finding that different convolution kernels are used to extract different features in the data~\cite{cnn2018overview}, we generate modules by removing the irrelevant convolution kernels in a CNN model. 
Compared to the original model, the module produced by compressed modularization has a different structure and fewer weights, resulting in less module reuse overhead than uncompressed modularization.
To decompose a trained CNN model for $N$-class classification, we formulate the modularization problem as a search problem. 
Search-based algorithms have been proven to be very successful in solving software engineering problems~\cite{harman2001search,li2016value}.
Given a space of candidate solutions, search-based approaches usually search for the optimal solution over the search space according to a user-defined objective function.
In the context of model decomposition, the candidate solution is defined as a set of sub-models containing a part of the CNN model's kernels, while the search objective is to search $N$ sub-models (as $N$ modules) with each of them recognizing one class.

\begin{figure}
    \centering
    \includegraphics[width=8.5cm]{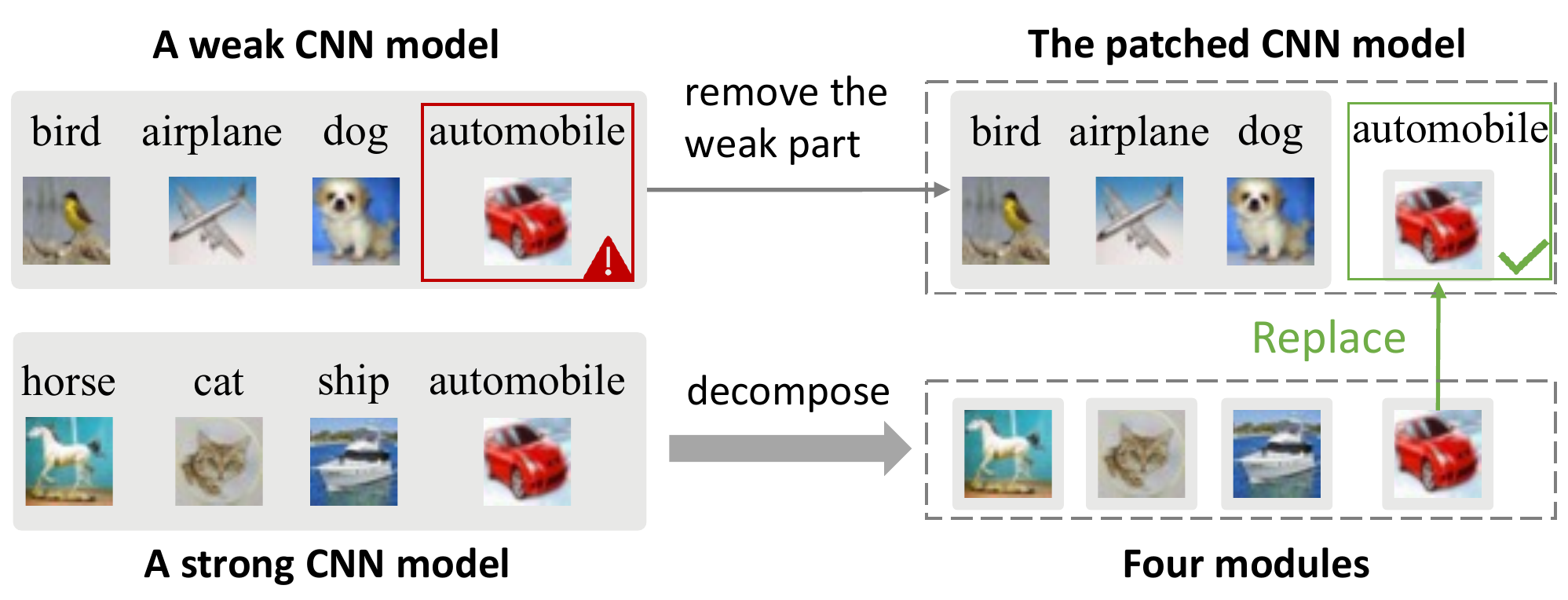}
    \caption{An example of using a module to patch a weak CNN model.}
    \label{fig:illustration}
    \vspace{-12pt}
\end{figure}

To patch a weak CNN model that achieves low performance on a target class (TC), as shown in Figure \ref{fig:illustration}, we compose the weak CNN model with the corresponding module obtained from a strong CNN model. 
The ability of the weak CNN model to recognize the TC can be improved through the patch,
as the patch (i.e., the patching module) comes from a strong model and can classify the TC better. 
During the prediction phase, the patched CNN model executes the weak CNN model and the patch in parallel. 
The prediction for the TC of the weak CNN model is replaced with that of the patch, resulting in the final prediction of the patched CNN model.
Note that, the model is defined as \textit{strong} when it preforms better on TC than the \textit{weak} model.
The weak model can be patched instead of being completely replaced by the strong model, since the weak model may outperform the strong model on some non-TCs and some non-TCs may not even be supported by the strong model. %

We evaluate \projectName using two representative CNN models with different structures (i.e., with and without residual connections ~\cite{resnet}) on three widely-used datasets (i.e., CIFAR-10 ~\cite{cifar10}, CIFAR-100~\cite{cifar10} and SVHN ~\cite{svhn}). 
The experimental results show that decomposing a trained CNN model into modules with \projectName and then composing the modules to build a new CNN model is functionally similar to the original one.
\projectName does not cause much loss of performance in terms of model accuracy (-2.89\% on average). 
In addition, each module retains 56.76\% of the convolution kernel of the trained CNN model on average. 
To validate the effectiveness of applying a module as a patch to improve weak CNN models, we conduct experiments for three common types of weak models, i.e., overly simple model, underfitting model~\cite{ma2018mode}, and overfitting model~\cite{overfitting,ma2018mode}. 
Overall, after patching, the averaged improvements on TC in terms of precision, recall, and F1-score are 12.54\%, 2.14\%, and 7.52\%, respectively.
Moreover, on non-TCs, 94\% of patched models outperform the weak models in terms of accuracy, with an average gain of 1.18\%. The detailed results and discussion are presented in Section~\ref{sec:experiments}.

The main contributions of this work are as follows:
\begin{itemize}[leftmargin=*]
    \item We propose an approach, named \projectName, to decomposing a CNN model into a set of reusable modules. We also apply \projectName to patch weak models without retraining.
    To our best knowledge, \projectName is the first \textit{compressed modularization} approach that 
    can decompose trained CNN models into CNN modules and reduce the overhead of module reuse.
    \item We formulate the modularization of CNNs as a search problem and design a genetic algorithm to solve it. %
    Especially, we propose the importance-based kernels grouping, the sensitivity-based initialization, and the pruning-based evaluation to alleviate the problem of excessive search space and time complexity in CNN modularization.
    \item We conduct extensive experiments using two representative CNN models on three widely-used datasets. The results show that \projectName can decompose a trained CNN model into modules without significant loss of model accuracy. Also, the experiments demonstrate the effectiveness of applying a module as a patch. 
\end{itemize}

\section{Background}
\label{sec:background}
\subsection{Convolutional Neuron Networks}
\label{subsec:cnn}
\begin{figure}[!h]
	\centering
	\includegraphics[width=8.5cm]{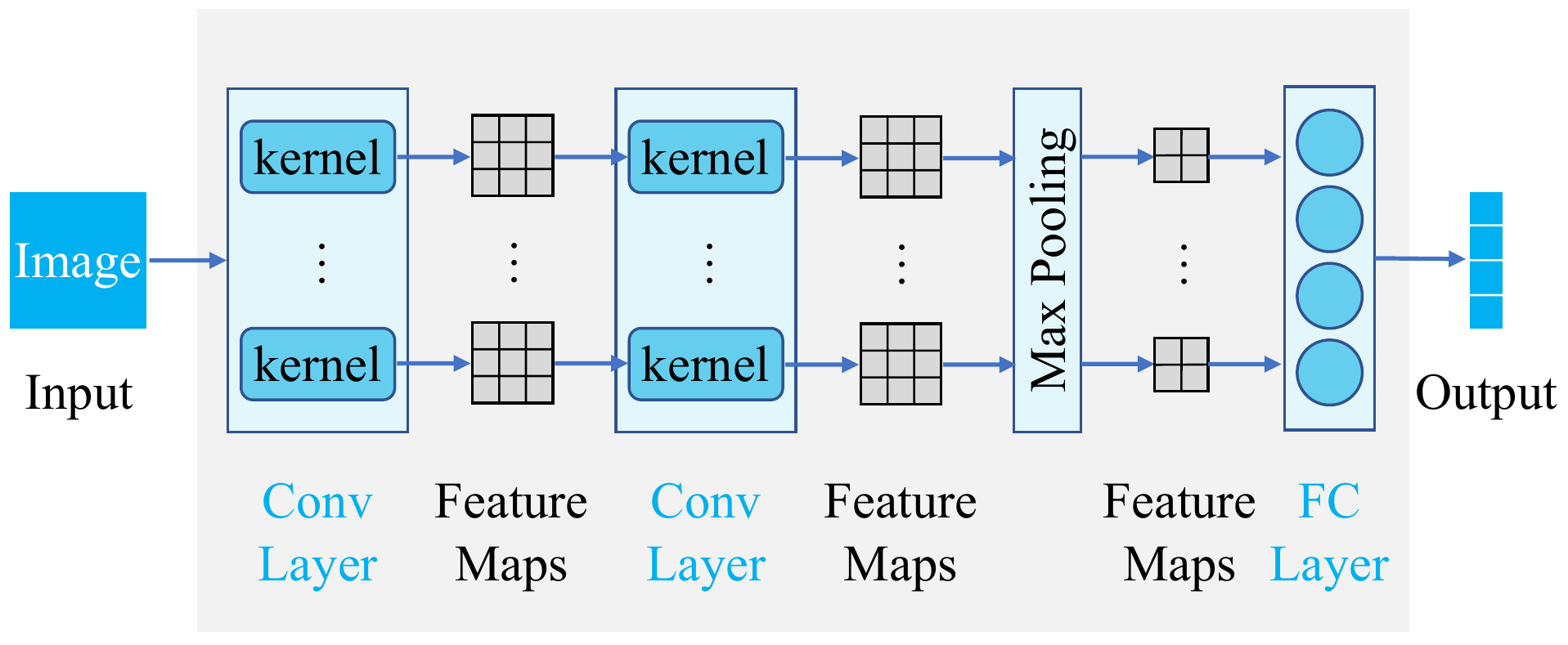}
    \caption{The architecture of a typical CNN model.}
    \label{fig:cnn-model}
    \vspace{-6pt}
\end{figure}

\begin{figure*}[t]
	\centering
	\includegraphics[width=17.5cm]{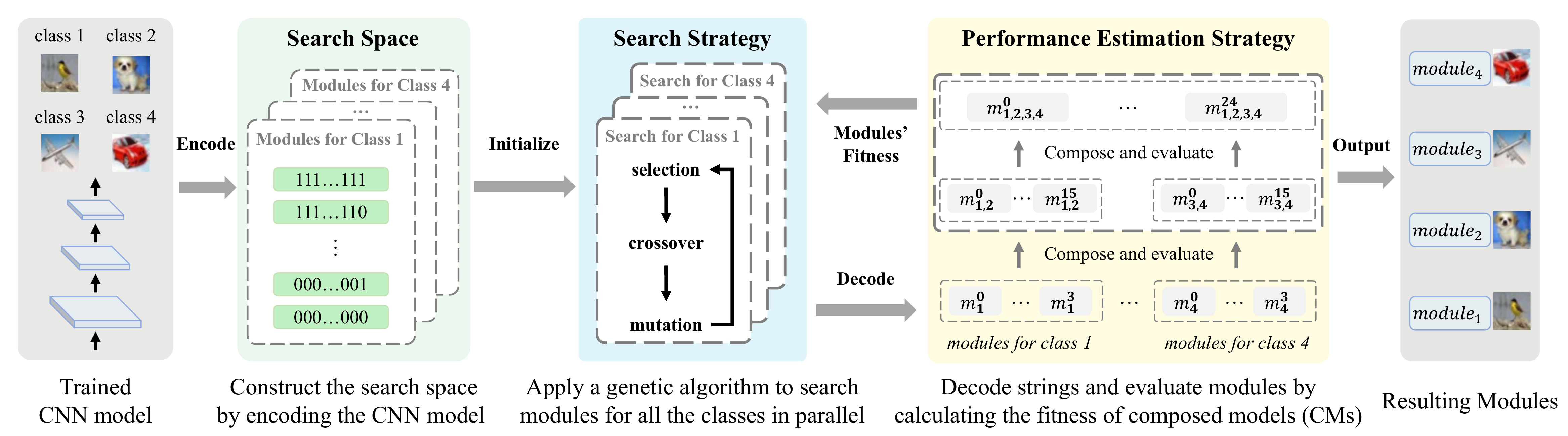}
	\vspace{-6pt}
    \caption{The overall workflow of \projectName.}
    \label{fig:framework}
    \vspace{-6pt}
\end{figure*}

\begin{figure}
    \centering
    \includegraphics[width=7.7cm]{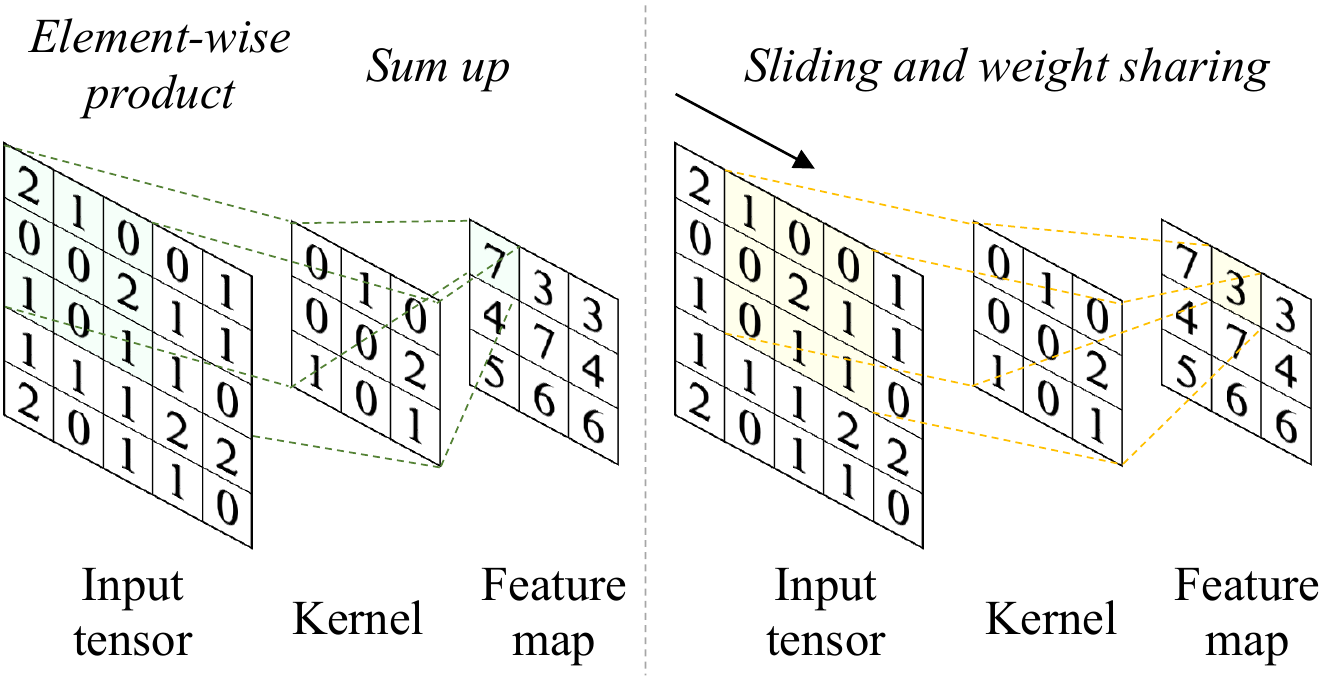}
    \vspace{-6pt}
    \caption{An example of convolution operation.}
    \label{fig:conv_oper}
    \vspace{-12pt}
\end{figure}

A typical CNN contains convolutional layers, pooling layers, and fully connected (FC) layers, of which the convolutional layers are the core of a CNN~\cite{cnn2018overview,vgg}. 
Figure \ref{fig:cnn-model} shows an overview of a simple CNN architecture. 
A convolutional layer contains many convolution kernels, each of which learns to recognize a local feature of an input tensor~\cite{cnn2018overview,lenet}. 
An input tensor can be the input image or a feature map produced by the previous convolutional layer or pooling layer. 
As shown in Figure \ref{fig:conv_oper}, by sliding over the input tensor, a convolution kernel calculates the convolution of input tensor and the kernel.
Since a convolution kernel slides over the input tensor to match features and produce a feature map, all values in the feature map share the same convolution kernel.
For instance, in the feature map of Figure~\ref{fig:conv_oper}, the values in the top-left (7) and top-middle (3) share the same kernel, i.e., weight.
Weight sharing~\cite{cnn2018overview,lenet} is one of the key features of a convolutional layer. 
The values in a feature map reflect the degree of matching between the kernel and the input tensor. For instance, compared to the position in the input tensor corresponding to the top-middle (3) in the feature map, the position corresponding to the top-left (7) is more similar to the kernel.
A pooling layer provides a sampling operation. 
The most popular form of pooling operation is max pooling, which reduces the dimensionality of the feature maps %
via downsampling operation.
FC layers are usually at the end of CNNs and are used to make predictions based on the features extracted from the convolutional and pooling layers.

\subsection{Genetic Algorithm}
\label{subsec:ga}
Inspired by the natural selection process, the genetic algorithm performs \textit{selection}, \textit{crossover}, and \textit{mutation} for several \textit{generations} (i.e., rounds) to generate solutions for a search problem~\cite{houck1995genetic, reeves1995genetic}.
A standard genetic algorithm has two prerequisites, i.e., the representation of an \textit{individual} and the calculation of an individual's \textit{fitness}.
For instance, the genetic algorithm is used to search for high-quality 
CNN architectures~\cite{evolution2017genetic, genetic2019}.
An \textit{individual} is a bit vector representing a NN architecture~\cite{evolution2017genetic}, where each bit corresponds to a convolution layer. 
The \textit{fitness} of an individual is the classification accuracy of a trained CNN model with the architecture represented by the individual. 
During the search, in each generation, the \textit{selection} operator compares the fitness of individuals and preserves the strong ones as parents that obtain high accuracy. 
The \textit{crossover} operator swaps part of two parents.
The \textit{mutation} operator randomly changes several bit values in the parents to enable or disable these convolution layers corresponding to the changed bit values. 
After the three operations, a new \textit{population} (i.e., a set of individuals) is generated. And the process continues with the new generation iteratively until it reaches a fixed number of generations or an individual with the target accuracy is obtained.

\section{Modularization of CNN Models}
\label{sec:approach}

Figure \ref{fig:framework} shows the overall workflow of \projectName. For a given trained $N$-class CNN model $\mathcal{M}{=}\{k_0, k_1,\dots, k_{L-1}\}$ with $L$ convolution kernels, the modularization process is summarized as follows: 

(1) \textit{Construction of Search Space}: \projectName encodes each candidate module into a fixed-length bit vector, 
where each bit represents whether the corresponding kernels are kept or not.
The bit vectors of all candidate modules constitute the search space.

(2) \textit{Search Strategy}: From the search space, the search strategy employs a genetic algorithm to find modules for $N$ classes.

(3) \textit{Performance Estimation}: The performance estimation strategy measures the performance (i.e., fitness) of the searched candidate and guides the search process. %

\subsection{Search Space}

As shown in Figure \ref{fig:framework}, the search space is represented using a set of bit vectors. %
For a CNN model with a lot of kernels, the size of vector could be very long, resulting in an excessively large search space, which could seriously impair the search efficiency. 
For instance, 10-class VGGNet-16~\cite{vgg} includes 4,224 kernels, so the number of candidate modules for each class is $2^{4224}$. 
In total, the size of the search space will be $10 \times 2^{4224}$. 
To reduce the search space, the kernels in a convolutional layer are divided into groups.
A simple way is to randomly group kernels; however,  
this could result in a group containing both kernels necessary for a module to recognize a specific class and those that are unnecessary. 
The randomness introduced by random grouping cannot be eliminated by subsequent searches, resulting in unnecessary kernels in the searched modules.

To avoid unnecessary kernels as much as possible, an \textit{importance-based grouping scheme} is proposed to group kernels based on their importance. 
As introduced in Section \ref{sec:background}, the values in a feature map can reflect the degree of matching between a convolution kernel and an input tensor. 
The kernels producing feature maps with weak activations are likely to be unimportant, as the values in the feature map with weak activations are generally small (and even zero) and have little effect on the subsequent calculations of the model~\cite{li2016pruning}. 
Inspired by this, \projectName measures the importance of kernels for each class based on the feature maps.
Specifically, given $m$ samples labeled class $n$ from the training dataset, a kernel outputs $m$ feature maps. %
We calculate the sum of all values in each feature map and use the average of $m$ sums to measure the importance of the kernel for class $n$. 
Then $L$ kernels are divided into $G$ groups following the importance order. 
Consequently, a module is encoded into a bit vector $[0,1]^G$, where each bit represents whether the corresponding group of kernels is removed. 
The number of candidate modules for the $n$-th class is $2^G$, and for $N$ classes, the search space size is reduced to $N \times 2^G$. %

For simplicity, if the number of kernels in a convolutional layer is less than 256, the kernels are divided into 10 groups; otherwise, they are divided into 100 groups. In this way, each kernel group has a moderate number of kernels (i.e., \~10) and groups in the same convolutional layer have approximately the same number of kernels. 

\subsection{Search Strategy}
\label{subsec:searchstrategy}
A genetic algorithm~\cite{evolution2017genetic} is used to search CNN modules, which has been widely used in search-based software engineering~\cite{geneticSE_1, geneticSE_2}.
The search process starts by initializing a population of $N_I$ individuals for each of $N$ classes. 
Then, \projectName performs $T$ generations, each of which consists of three operations (i.e., selection, crossover, and mutation) and produces $N_I$ new individuals for each class. 
The fitness of individuals is evaluated via a performance estimation strategy that will be introduced in Section \ref{subsec:performance}. 

\subsubsection{Sensitivity-based Initialization} 
In the $0$-th generation, a set of modules $M_n^0=\{m_{n, i}^0\}_{i=0}^{N_I-1}$ are initialized for class $n$,
where $n=0, 1, \dots, N-1$ and $m_{n, i}^0$ is a bit vector $[0,1]^G$. %
Two schemes are used to set the bits in each individual (i.e., module): random initialization and \textit{sensitivity-based initialization}. 
Random initialization is a common scheme~\cite{nas2019survey, evolution2017genetic}. Each bit in an individual is independently sampled from a Bernoulli distribution. %
However, random initialization causes the search process to be slow or even fail (see Section \ref{subsec:result}). 
We observed a phenomenon that some convolutional layers are sensitive to the removal of kernels, which has been also observed in network pruning ~\cite{li2016pruning}. That is, the accuracy of a CNN model dramatically drops when some particular kernels are dropped from a sensitive convolutional layer, while the loss of accuracy is not more than 0.01 when many other kernels (e.g., 90\% of kernels) are dropped from an insensitive layer.

To evaluate the sensitivity of each convolutional layer, we drop out 10\% to 90\% kernels in each layer incrementally and evaluate the accuracy of the resulting model on the validation dataset. 
If the loss of accuracy is small (e.g., within 0.05) when 90\% kernels in a convolutional layer are dropped, the layer is insensitive, otherwise, it is sensitive.
When initializing $m_{n, i}^0$ using sensitivity-based initialization, fewer kernel groups are dropped from the sensitive layers while more kernel groups from the insensitive layers. 
More specifically, 
a drop ratio is randomly selected from 10\% to 50\% for a sensitive layer (i.e., 10\% to 50\% bit values are randomly set to 0). 
In contrast, a drop ratio is randomly selected from 50\% to 90\% for an insensitive layer.

\subsubsection{Selection, Crossover, and Mutation}
\label{subsubsec:selection}
For class $n$, to generate the population (i.e., modules) of the $t$-th generation, \projectName performs selection, crossover, and mutation operations on the $(t-1)$-th generation's population $M_n^{t-1}$.
First, the selection operation selects $N_P$ individuals from $M_n^{t-1}$ as parents according to individuals' fitness. 
Then, the single-point crossover operation generates two new individuals by exchanging part of two randomly chosen parents from $N_P$ parents. 
Next, the crossover operation iterates until $N_I$ new individuals are produced. 
Finally, the mutation operation on the $N_I$ new individuals involves flipping each bit independently with a probability $p_M$. 
For $N$ classes, selection, crossover, and mutation operations are performed in parallel, resulting in a total of $N \times N_I$ modules. 

\subsection{Performance Estimation Strategy}
\label{subsec:performance}
A module with high fitness should have the same good identification ability as the trained model $\mathcal{M}$ and only recognize the features of one specific class. 
Two evaluation metrics are used to evaluates the fitness of modules: the \textit{accuracy} and the \textit{difference} of modules.
The higher the \textit{accuracy}, the stronger the ability of the module to recognize features of the specific class. 
On the other hand, 
the greater the \textit{difference}, the more a module focuses on the specific class. 
In addition, the \textit{difference} can be used as a regularization to prevent the search from overfitting the \textit{accuracy}, as the simplest way to improve \textit{accuracy} is to allow each module to retain all the convolution kernels of $\mathcal{M}$. 
Consequently, the fitness of a module is the weighted sum of the \textit{accuracy} and the \textit{difference}. 
Furthermore, when calculating the fitness, a pruning strategy is used to improve the evaluation efficiency, making the performance estimation strategy computationally feasible.

\subsubsection{Evaluation Metrics}
\label{subsubsec:metrics}
Since a module focuses on a specific class and is equivalent to a single-class classifier, we combine modules into a composed model (CM) to evaluate them. 
That is, one module is selected from each class's $N_I$ modules, and the $N$ modules are combined into a $CM^{(N)}$ for $N$-class classification.
The $CM^{(N)}$ is evaluated on the same classification task as $\mathcal{M}$ using the dataset $D$.
The accuracy of $CM^{(N)}$ and the difference between the modules within $CM^{(N)}$ are assigned to each module.
Specifically, the accuracy and difference of each module are calculated as follows: 

\textbf{\textit{Accuracy (Acc). }}%
To calculate the \textit{Acc} of $CM^{(N)}$, the $N$ modules are executed in parallel, and the output of $CM^{(N)}$ is obtained by combining modules' outputs. 
Specifically, given a $CM^{(N)}{=}\{m_n\}_{n=0}^{N-1}$, the output of module $m_n$ for an input $j$ labeled $class_j$ is a vector $O_{n,j}{=}[o_n^0, o_n^1, \dots, o_n^{N-1}]$, where each value corresponds to a class. Since $m_n$ is used to recognize class $n$, the $n$-th value $o_n^n$ is retained. 
Consequently, the output of $CM^{(N)}$ is $O_j{=}[o_0^0, o_1^1 \dots, o_{N-1}^{N-1}]$, and the \textit{Acc} of $CM^{(N)}$ is calculated as follows:
\begin{gather}
    Acc = \frac{1}{|D|} \sum_j^{|D|} pred(j), \\
    pred(j) = 
    \begin{cases}
    1, & \mbox{if }\mathop{\arg\max}\limits_{n=0, 1, \dots, N-1}O_j = class_j \\
    0, & \mbox{if }\mathop{\arg\max}\limits_{n=0, 1, \dots, N-1}O_j \ne class_j .
    \end{cases}
\end{gather} 

\textbf{\textit{Difference (Diff). }}%
Since a module can be regarded as a set of convolution kernels, the difference between two modules can be measured by the Jaccard Distance (\textit{JD}) that measures the dissimilarity between two sets. 
The \textit{JD} between set $A$ and set $B$ is obtained by dividing the difference of the sizes of the union and the intersection of two sets by the size of the union: 
\begin{gather}
    JD(A, B) = \frac{|A \cup B| - |A \cap B|}{|A \cup B|}. \label{eq:jaccard}
\end{gather}
If the $JD(A,B)=1$, there is no commonality between set $A$ and set $B$, and if it is 0, then they are exactly the same.
Based on \textit{JD}, the \textit{Diff} value of $CM^{(N)}$ is the average value of \textit{JD} between all modules:
\begin{gather}
    Diff = \frac{2}{N \times (N-1)} \times \sum_{0 \leq i < j \leq N-1} JD(m_i, m_j). \label{eq:diff}
\end{gather}
Based on \textit{Acc} and \textit{Diff}, the fitness value of $CM^{(N)}$ is calculated via:
\begin{gather}
    fitness = \alpha \times Acc + (1-\alpha) \times Diff, \label{eq:fitness}
\end{gather}
where $\alpha$ is a weighting factor and $0 < \alpha < 1$. In practice, $\alpha$ is set to a high value (e.g., 0.9) because high accuracy is a prerequisite for the availability of modules.
The fitness value of $CM^{(N)}$ is then assigned to the $N$ modules within $CM^{(N)}$. 
Since each module is used in multiple CMs, a set of fitness values is assigned to each module. 
The maximum value of the set is a module's final fitness. 

\subsubsection{Decode}
\begin{figure}
    \centering
    \includegraphics[width=8cm]{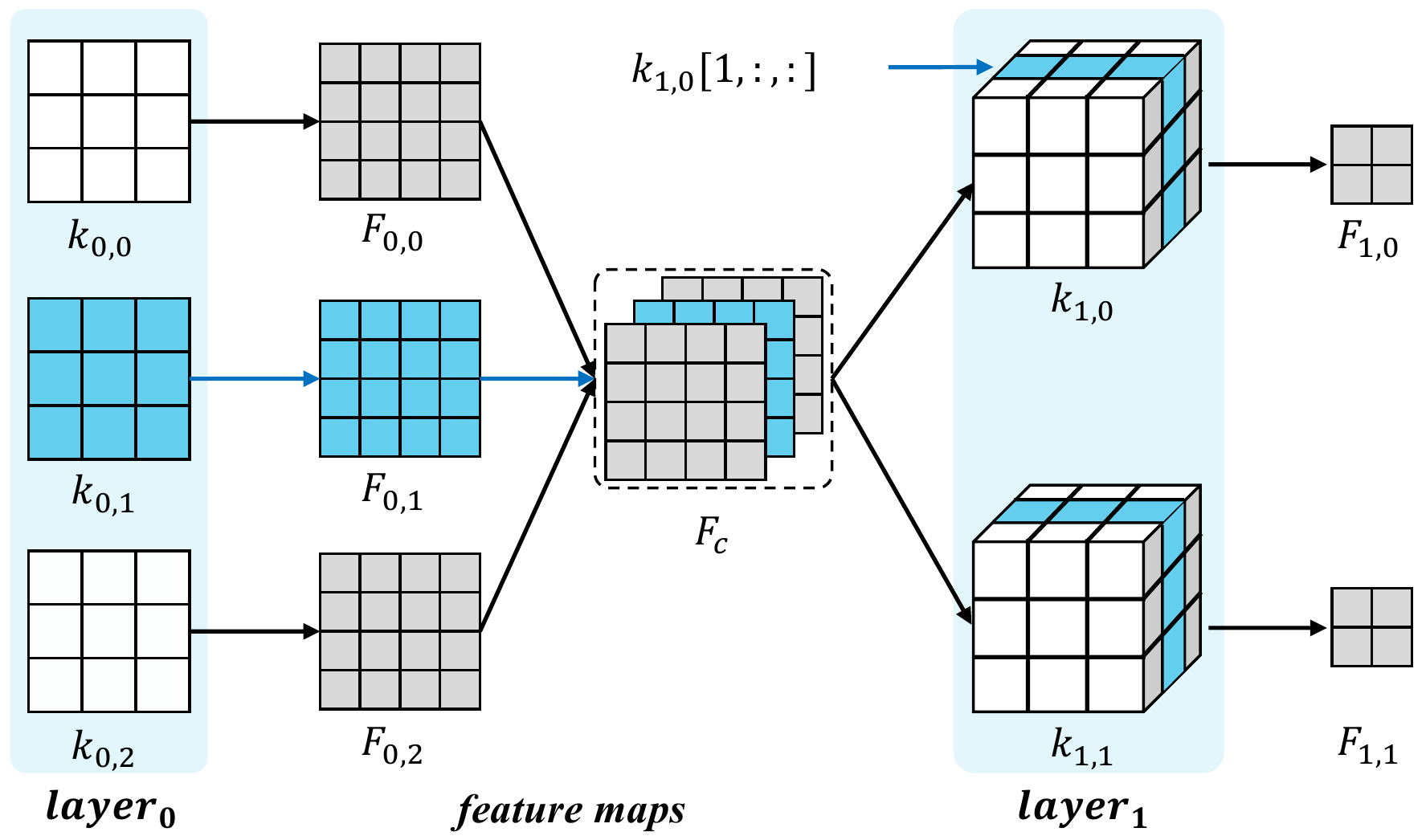}
    \caption{The process of removing convolution kernels.}
    \label{fig:decode}
    \vspace{-12pt}
\end{figure}

To evaluate modules, each bit vector is transformed into a runnable module by removing the kernel groups from $\mathcal{M}$ corresponding to the bits of value 0.
Since removing kernels from a convolutional layer affects the convolutional operation in the later convolutional layer, the kernels in the latter convolutional layer need to be modified to ensure that the module is runnable. 

Figure \ref{fig:decode} shows the process of removing convolution kernels. 
During the convolution, the three kernels $k_{0,*} \in \mathbb{R}^{3 \times 3}$ in $layer_0$ output three feature maps $F_{0,*} \in \mathbb{R}^{4 \times 4}$ that are then combined in a feature map $F_c \in \mathbb{R}^{3 \times 4 \times 4}$ and fed to $layer_1$. 
By sliding on $F_c$, kernels $k_{1,*} \in \mathbb{R}^{3 \times 3 \times 3}$ in $layer_1$ perform the convolution and output two feature maps $F_{1,*} \in \mathbb{R}^{2 \times 2}$. 
If $k_{0,1}$ in $layer_0$ is removed, the feature map $F_{0,1}$ generated by $k_{0,1}$ is also removed. 
The input of $layer_1$ becomes a different feature map $F^{'}_c \in \mathbb{R}^{2 \times 4 \times 4}$, the dimension of which does not match that of $k_{1,*} \in \mathbb{R}^{3 \times 3 \times 3}$ in $layer_1$, causing the convolution to fail. 

To solve the dimension mismatch problem, we remove the part of $k_{1,*}$ that corresponds to $F_{0,1}$, ensuring the first dimension of $k_{1,*}$ to match with that of $F^{'}_c$. 
For instance, since $F_{0,1}$ is removed, $k_{1,0}[1,:,:]$, which performs convolution on $F_{0,1}$, becomes redundant and causes the dimension mismatch. 
We remove $k_{1,0}[1,:,:]$ and the transformed kernel $k^{'}_{1,0} \in \mathbb{R}^{2 \times 3 \times 3}$ can perform convolution on $F^{'}_c$. 

In addition, since the residual connection adds up the feature maps output by two convolutional layers, the number of kernels removed from the two convolutional layers must be the same to ensure that the output feature maps match in dimension. 
When constructing a bit vector, we treat the two convolutional layers as one layer and use the same segment to represent the two layers so that they always remove the same number of kernels. 

\subsubsection{Pruning-based Evaluation}

Since the fitness of each module comes from the one with the highest fitness among the CMs the module participates in, the number of $CM^{(N)}$ that \projectName needs to evaluate is $(N_I)^N$.
The time complexity is $O(n^N)$, which could be too high to finish the evaluation in a limited time. 
To reduce the overhead, a pruning strategy is designed, which is based on the following fact: if the accuracy of $CM^{(N)}$ is high, the accuracy of the $CM^{(n)}$ (e.g., $CM^{(2)}$ for the binary classification) composed of the modules within the $CM^{(N)}$ is also high. 
If the accuracy of a module is low, the accuracy of the $CM^{(n)}$ containing the module is also lower than the $CM^{(n)}$ containing modules with high accuracy.
In addition, the number of $CM^{(n)}$ is much smaller than that of $CM^{(N)}$. 
For instance, the $N$-class classification task can be decomposed into $N/2$ binary classification subtasks, resulting in $N/2 \times (N_I)^2$ $CM^{(2)}$. 

Therefore, the $N$-class classification task is decomposed into several subtasks. The accuracy of $CM^{(n)}$ is evaluated, and the top $N_{top}$ $CM^{(n)}$ with high accuracy are selected to be combined into $CM^{(N)}$.
Through continuous evaluation, selection, and composition, a total of $(N_{top})^2$ $CM^{(N)}$ are composed. 
The time complexity is $O(n^2)$, which is lower than the original time complexity $O(n^N)$. 

\section{Patching Weak CNN Models through Modularization and Composition}
\label{subsec:patching}

\begin{figure}[!t]
    \centering
    \includegraphics[width=8cm]{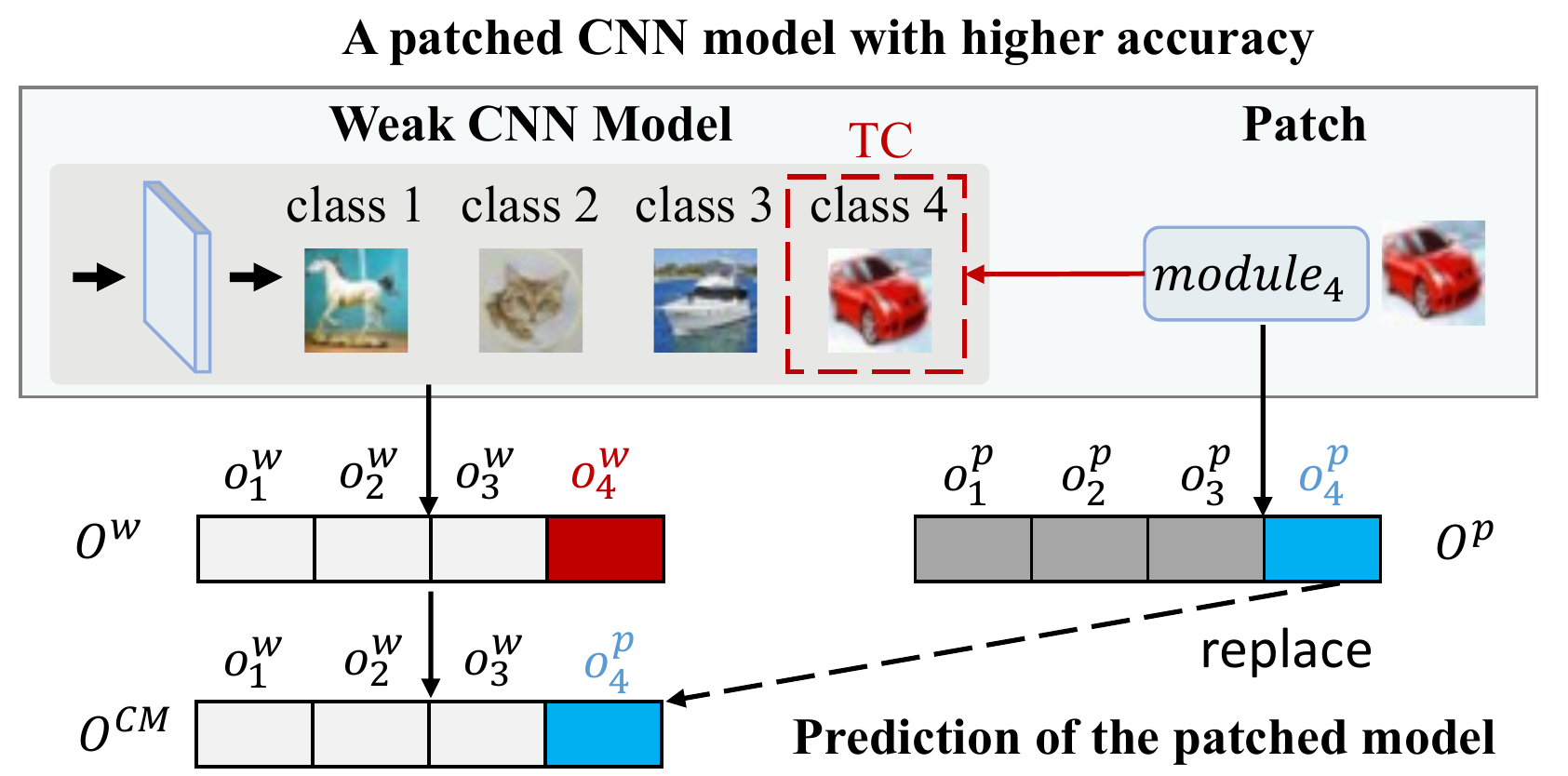}
    \caption{Patching a weak CNN model.}
    \label{fig:patching}
    \vspace{-12pt}
\end{figure}

The weak CNN model can be improved by patching the target class (TC).
To identify the TC of a weak CNN model, developers can use test data to evaluate the weak CNN model's classification performance (e.g., precision and recall) of each class. The class in which the weak CNN model achieves poor classification performance is regarded as TC. 
As illustrated in Figure \ref{fig:patching}, 
the TC is replaced with the corresponding module from a strong model. 
To find the corresponding module, a developer can evaluate the accuracy of a candidate model on TC. 
If the candidate model's accuracy exceeds that of the weak model, it can be used as a patch. 

Formally, 
given a weak CNN model $\mathcal{M}_w$, suppose there exists a strong CNN model $\mathcal{M}_s$ whose classification task intersects with that of $\mathcal{M}_w$. For instance, both $\mathcal{M}_w$ and $\mathcal{M}_s$ can recognize TC $n$. Then, the corresponding module $m_n$ from $\mathcal{M}_s$ can be used as a patch to improve the ability of $\mathcal{M}_w$ to recognize TC $n$. 
Specifically, $\mathcal{M}_w$ and $m_n$ are composed into a CM that is the patched CNN model. 
Given an input, $\mathcal{M}_w$ and $m_n$ run in parallel and the outputs of them are $O^w=[o^w_0, o^w_1, \dots, o^w_{N-1}]$ and $O^p=[o^p_0, o^p_1, \dots, o^p_{N-1}]$, respectively. 
Then, the output $O^{CM}$ of CM is obtained by replacing the prediction corresponding to TC $n$ of $O^w$ with that of $O^p$. 

A straightforward way is to directly replace $o^w_n$ with $o^p_n$, and the index of the maximum value in $O^{CM}=[o^w_0, \dots, o^p_n, \dots, o^w_{N-1}]$ is the predicted class. 
However, the comparison between $o^p_n$ and the other values in $O^{CM}$ is problematic: since $\mathcal{M}_w$ and $\mathcal{M}_s$ are different models that are trained on the different datasets or have different network structures, there could be significant differences in the distribution between the outputs of $\mathcal{M}_w$ and $\mathcal{M}_s$. 
For instance, we have observed that the output values of a model could be always greater than that of the other one, resulting in the outputs of a module decomposed from the strong model being always larger/smaller than the outputs of a weak model. 
This problem could cause error prediction when calculating the prediction of CM; thus, $O^w$ and $O^p$ are normalized before the replacement. 
Specifically, since the outputs on the training set can reflect the output distribution of a module, we collect the outputs of $m_n$ on the training data with the class label $n$. 
Then, the minimum and maximum values of the output's distribution can be estimated using the collected outputs.  
For instance, ($min$, $max$) are the minimum and maximum values of the collected outputs of $m_n$. 
The normalized $o^p_n$ is $\frac{o^p_n-min}{max - min}$. 
In addition, the $softmax$ is used over $O^w$ to scale the values in $O^w$ between 0 and 1. 
Finally, the prediction of CM is obtained by replace $o^w_n$ in normalized $O^w$ with normalized $o^p_n$.

\section{Experiments}
\label{sec:experiments}

\subsection{Research Questions}
We evaluate \projectName to answer following research questions:
\begin{itemize}[leftmargin=*]
\item \textbf{RQ1: How effective is \projectName in modularizing CNN models?} Modularization should ensure that modules retain sufficient ability (i.e., sufficient convolution kernels) to recognize features. 
On the other hand, compared to the trained CNN model, the smaller the module (i.e., the fewer convolution kernels), the lower the memory and computational overhead from the patch. 
In this research question, the modularization results of four models are used to validate whether \projectName can strike a balance between the module's ability and its size. 

\item \textbf{RQ2: Can weak CNN models be improved through modularization and composition?} 
In this research question, we conduct experiments for three common types of weak CNN models, i.e., overly simple, underfitting, and overfitting models, to validate the effectiveness of applying a module as a patch to improve a weak CNN model.  

\item \textbf{RQ3: How effective are the three heuristic methods in \projectName?}
{As described in Section \ref{sec:approach}, we propose three heuristic methods in the design of \projectName, namely importance-based grouping, sensitivity-based initialization, and pruning-based evaluation. In this RQ, we evaluate the effectiveness of these methods.}

\end{itemize}

\subsection{Dataset and Models}
\label{subsec:benchmarks}

The following three datasets are used to evaluate \projectName, which are also widely used in related research~\cite{nnmodularity2022icse,feng2020deepgini} on deep learning testing.

\textbf{CIFAR-10.} The CIFAR-10~\cite{cifar10} dataset is used to train strong CNN models. CIFAR-10 contains natural images with resolution $32\times32$, which are drawn from 10 classes including airplanes, cars, birds, cats, deer, dogs, frogs, horses, ships, and trucks. 
The training and test sets contain 50,000 and 10,000 images respectively. 
For CIFAR-10, 10,000 images are selected from the training set to form the validation set. 

\textbf{CIFAR-100.} The CIFAR-100~\cite{cifar10} dataset is used to train weak models. CIFAR-100 consists of $32\times32$ natural images in 100 classes, with 600 images per class. There are 500 training images and 100 testing images per class.  
To reduce the time overhead required to train weak models, a subset of CIFAR-100 is built, consisting of 9 classes: apple, baby, bed, bicycle, bottle, bridge, camel, clock, and rose. 
Each class in CIFAR-10 is considered as TC in turn and merged with the subset, resulting in ten 10-class classification datasets.
The ten datasets are used to train weak models.
The proportion for training, validation, and testing data is 8:1:1.

\textbf{SVHN.} The Street View House Number (SVHN) dataset ~\cite{svhn} contains colored digit images 0 to 9 with resolution $32\times32$. 
All the 604,388 training images are used, 20\% of them are chosen for validation. The test set contains 26,032 images. 
A subset of SVHN, namely SVHN-S, includes images of $class_s=\{0, 1, 2, 3 ,4\}$ and is used to train strong models.
Five subsets of SVHN, namely SVHN-W, are used to train weak models, and each subset includes images of $class_w=\{6, 7, 8, 9, TC\}$, where $TC \in class_s$. 

The following two CNN models are used to evaluate \projectName:

\textbf{SimCNN} is constructed by stacking convolutional layers, which represents a basic structure CNN and is similar to LeNet~\cite{lenet}, AlexNet~\cite{alexnet}, and VGGNet~\cite{vgg}. 
The output of each convolutional layer can only flow through each layer in sequential order.
Without loss of generality, SimCNN in our experiments is set to contain 13 convolutional layers and 3 FC layers, totally 4,224 convolution kernels.

\textbf{ResCNN} is constructed by convolutional layers and residual connections, which represents a complex structure of CNNs and is similar to ResNet~\cite{resnet}, WRN~\cite{wrn}, and MobileNetV2~\cite{mobilenetv2}. 
A residual connection can go across one or more convolutional layers, allowing the output of a layer not only to flow through each layer in sequential order, but also to be able to connect with any following layer.
Without loss of generality, ResCNN in our experiments is set as 12 convolutional layers, 1 FC layer, and 3 residual connections, totally 4,288 convolution kernels.

All the experiments are conducted on Ubuntu 20.04 server with 64 cores of 2.3GHz CPU, 128GB RAM and NVIDIA Ampere A100 GPUs with 40 GB memory.

\subsection{Experimental Results}
\label{subsec:result}
\begin{table}[]
\caption{The modularization results of \projectName.
}
\vspace{-6pt}
\label{tab:RQ1}
\begin{center}
{
\resizebox{\columnwidth}{!}{
\begin{tabular}{cccccc}
\hline
\multirow{2}{*}{\textbf{Model Name}} & \multicolumn{2}{c}{\textbf{Trained Model}} & \multicolumn{2}{c}{\textbf{Composed Model}} & \multirow{2}{*}{\textbf{\begin{tabular}[c]{@{}c@{}}Avg. \# Kernels \\ in a Module\end{tabular}}} \\ \cmidrule(lr){2-3} \cmidrule(lr){4-5}
                                     & \textbf{Acc}     & \textbf{\# Kernels}     & \textbf{Acc}         & \textbf{Diff}        &                                                                                                  \\ \hline \hline
SimCNN-CIFAR                         & 0.8977           & 4224                    & 0.8607               & 0.5277               & 2617                                                                                             \\ \hline
SimCNN-SVHN                          & 0.9541           & 4224                    & 0.9385               & 0.6161               & 2230                                                                                             \\ \hline
ResCNN-CIFAR                         & 0.9041           & 4288                    & 0.8564               & 0.5648               & 2498                                                                                             \\ \hline
ResCNN-SVHN                          & 0.9506           & 4288                    & 0.9352               & 0.6046               & 2317                                                                                             \\ \hline
\end{tabular}
}
}
\end{center}
\end{table}

\begin{figure}
	\centering
	\includegraphics[width=\columnwidth]{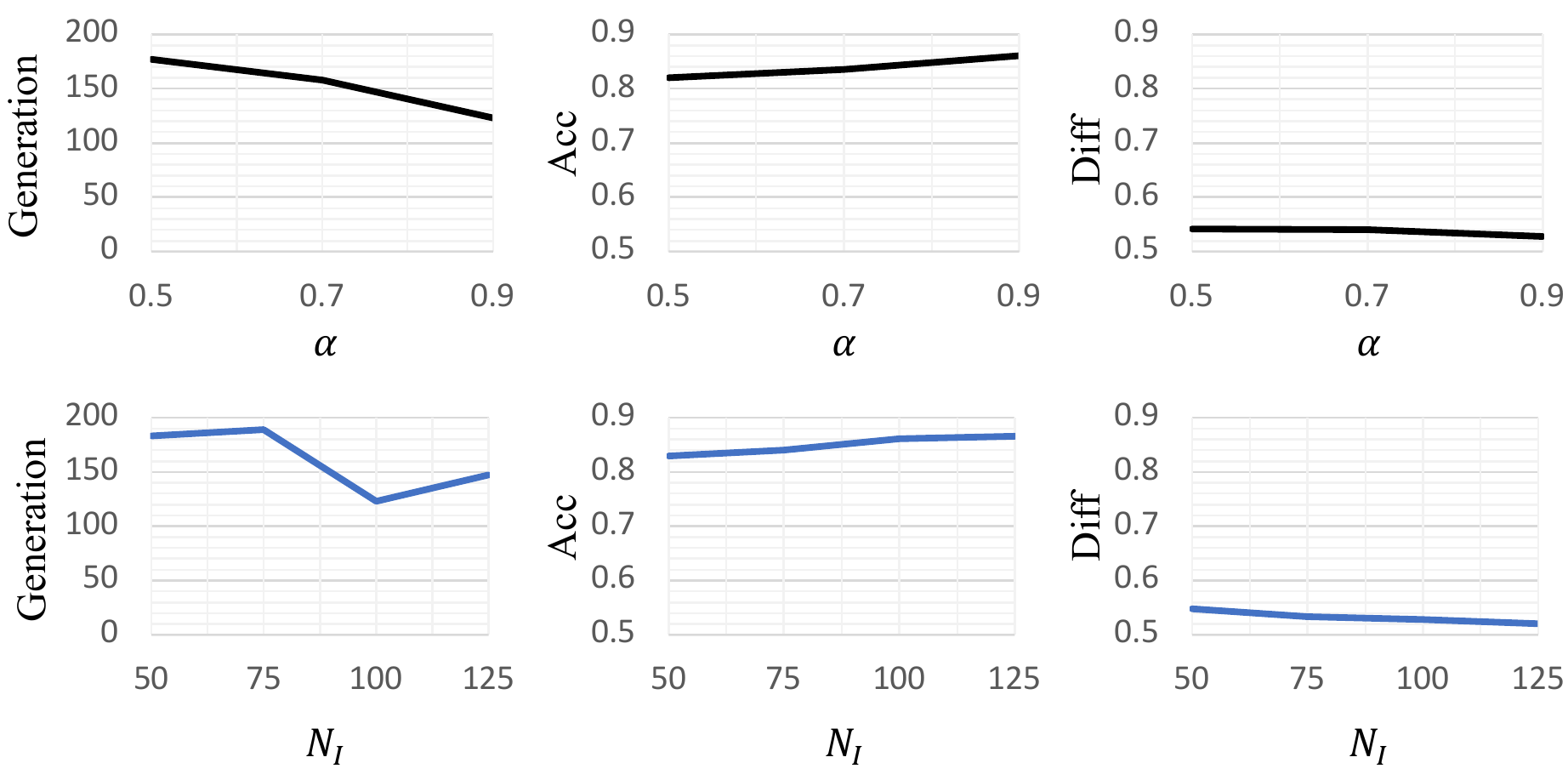}
    \caption{The impact of major parameters.}
    \label{fig:impact}
    \vspace{-12pt}
\end{figure}

\noindent\textbf{\textit{RQ1: How effective is \projectName in modularizing CNN models?}}

To answer RQ1, four CNN models are trained from scratch and modularized by \projectName. 
Specifically, SimCNN and ResCNN are trained on CIFAR-10 and SVHN-S from scratch, resulting in four strong CNN models: SimCNN-CIFAR, SimCNN-SVHN, ResCNN-CIFAR, and ResCNN-SVHN.
On both CIFAR-10 and SVHN-S datasets, SimCNN and ResCNN are trained with mini-batch size 128 for 200 epochs.  
The initial learning rate is set to 0.01 and 0.1 for SimCNN and ResCNN, respectively, and the initial learning rate is divided by 10 at the 60-th and 120-th epoch for SimCNN and ResCNN respectively. 
All the models are trained using data augmentation~\cite{dataaugmentation} and SGD with a weight decay~\cite{weightdecay} of $10^{-4}$ and a Nesterov momentum ~\cite{sutskever2013importance} of 0.9.

The four trained CNN models are then modularized by \projectName. 
Specifically, when applying genetic algorithm to search CNN modules, following the common practice~\cite{genetic2019, suganuma2018exploiting, real2017large}, the number of individuals $N_I$ and the number of parents $N_P$ in each generation are set to 100 and 50, respectively.
The mutation probability $p_M$ is generally small~\cite{genetic2019, suganuma2018exploiting} and is set to $0.1$. 
The weighting factor $\alpha$ is set to $0.9$. 
For the sake of time, an early stopping strategy ~\cite{q-learning-earlystopping, goodfellow2016deep} is applied, and the maximum number of generations is set as $T=200$. 
A trained CNN model $\mathcal{M}$ is modularized with reference to the validation set, which was not used in model training. 
After completing the modularization, the resulting modules are evaluated on the testing set.

When evaluating the resulting modules, the modules are first composed to construct a new composed model $CM$ that is functionally equivalent to $\mathcal{M}$ (e.g., 10-class classification model for CIFAR-10).
Then the \textit{Acc} of $\mathcal{M}$ and the \textit{Acc} of $CM$ are compared to evaluate the ability of modules to recognize features. 
The closer the \textit{Acc} of $CM$ is to the \textit{Acc} of $\mathcal{M}$, the better the ability of the modules in $CM$ to recognize features is preserved. 
\textit{Diff} is calculated to verify whether there are differences between the modules for different features. 
The closer the \textit{Diff} is to 1, the greater the difference between the modules. 
Furthermore, we calculate the number of convolution kernels retained by each module. %
The fewer kernels are retained, the less memory and computational overhead the patch brings.

Table \ref{tab:RQ1} shows the modularization results of \projectName on the four trained CNN models: SimCNN-CIFAR, SimCNN-SVHN, ResCNN-CIFAR, and ResCNN-SVHN.
The \textit{Acc} and \textit{Diff} of the composed CNN models are (0.8607, 0.5277), (0.9385, 0.6161), (0.8564, 0.5648), and (0.9352, 0.6046), respectively. 
Compared to the accuracy of the trained CNN models, the loss of accuracy caused by modularization is 0.037, 0.0156, 0.0477, and 0.0154, respectively. 
The average loss of accuracy is 0.0289, which demonstrates that the composed CNN models without retraining have comparable accuracy to the trained CNN models. 
The comparable accuracy of the composed CNN models suggests that the modules have sufficient ability to recognize features. 
The averaged \textit{Diff} is 0.5783, which means that there are indeed differences among the modules. 
We also count the number of convolution kernels in the modules for each model.
For SimCNN-CIFAR, SimCNN-SVHN, ResCNN-CIFAR, and ResCNN-SVHN, the average number of modules' kernels is 2617, 2230, 2498, and 2317, respectively. 
The modules of the four models respectively drop 38.04\%, 47.21\%, 41.74\%, and 45.97\% kernels from the original models.
On average, the number of kernels of a module is reduced by 43.24\%, indicating that the memory and computation costs of the module are smaller than the original model.

\begin{table}[t]
\caption{The FLOPs of the original model and decomposed modules}
\label{tab:flops}
\vspace{-6pt}
\resizebox{0.9\columnwidth}{!}{

\begin{tabular}{cccc}
\hline
\textbf{Model Name} & \textbf{\begin{tabular}[c]{@{}c@{}}Model\\ FLOPs (M)\end{tabular}} & \textbf{\begin{tabular}[c]{@{}c@{}}Module\\ FLOPs (M)\end{tabular}} & \textbf{Reduction} \\ \hline \hline
SimCNN-CIFAR        & \multirow{2}{*}{313.7}                                             & 164.3                                                               & 47.6\%             \\
SimCNN-SVHN         &                                                                    & 107.8                                                               & 65.6\%             \\ \hline
ResCNN-CIFAR        & \multirow{2}{*}{431.2}                                             & 225.4                                                               & 47.7\%             \\
ResCNN-SVHN         &                                                                    & 142.1                                                               & 67.0\%             \\ \hline
\multicolumn{3}{c}{\textbf{Average}}                                                                                                                           & \textbf{57.0\%}    \\ \hline
\end{tabular}

}
\end{table}

The memory and computation costs of modules are important for DNN modularization, which %
incur module reuse overhead. 
The memory and computation costs can be measured by the number of weights (i.e., the number of kernels in Table \ref{tab:RQ1}) and the floating point operations (FLOPs)~\cite{li2016pruning}, respectively.
In this experiment, an open-source tool fvcore~\cite{fvcore} is used to calculate the FLOPs required by the modules or models to classify an image.
As shown in Table~\ref{tab:flops}, for our approach, SimCNN-* and ResCNN-* require 313.7 million and 431.2 million FLOPs, respectively.
The third column shows the averaged FLOPs of modules, and the last column shows the percentage of reduction in FLOPs over the original model.
For instance, the average FLOPs required by the modules of SimCNN-CIFAR are 164.3 million, which are 47.6\% lower than the FLOPs of the originally trained model.
On average, the FLOPs required by a module are 57.0\% lower than the original model. %
In contrast, uncompressed modularization approaches~\cite{fse2020modularity,nnmodularity2022icse} incur higher module reuse overhead. 
We analyze the open source projects~\cite{moduleToolA, moduleToolB} published by ~\cite{fse2020modularity,nnmodularity2022icse}, including source code files and the experimental data (e.g., the trained CNN models and the generated modules). 
The open-source tool keras-flops~\cite{kerasFlops} is used to calculate the FLOPs for the approach described in ~\cite{fse2020modularity}. 
The FLOPs required by a module in ~\cite{fse2020modularity} are the same as those required by the original model.
For the project of~\cite{nnmodularity2022icse}, since the modules are not encapsulated as Keras model, there are no ready-to-use, off-the-shelf tools to calculate the FLOPs required by the modules. We manually analyze the number of weights of the module and confirm that a module has the same number of weights as the original model.
In summary, the experimental results indicate that \projectName incurs less module reuse overhead than the uncompressed modularization approaches~\cite{fse2020modularity,nnmodularity2022icse}.

In addition, we investigate the impact of major parameters on \projectName, including $\alpha$ (the weighting factor between the \textit{Acc} and the \textit{Diff}, described in Sec. \ref{subsec:performance}) and $N_I$ (the number of modules in each generation, described in Sec. \ref{subsec:searchstrategy}). 
Figure \ref{fig:impact} shows the ``Generation'', ``Acc'', and ``Diff'' of the composed model on SimCNN-CIFAR with different $\alpha$ and $N_I$. %
We find that, \projectName performs stably under different parameter settings in terms of \textit{Acc} and \textit{Diff}. The changes in the number of generations show that a proper setting can improve the efficiency of \projectName. 
The results also show that our default settings (i.e., $\alpha=0.9$ and $N_I=100$) are appropriate.

\begin{tcolorbox}
\vspace{-6pt}
The average loss of accuracy caused by modularization is 0.0289.
The number of kernels and FLOPs of a module are reduced by 43.24\% and 57.0\%, respectively.
Experimental results demonstrate that \projectName strikes a balance between the module's ability and its reuse overhead. 
\vspace{-6pt}
\end{tcolorbox}

\ 

\noindent\textbf{\textit{RQ2: Can weak CNN models be improved through modularization and composition?}}
\begin{table}[]
\caption{Precision of weak models and patched models on CIFAR. All results are in \%.}
\label{tab:patch_cifar_precision}
\vspace{-6pt}
\resizebox{\columnwidth}{!}{
\begin{tabular}{ccrrrrrr}
\hline
\multirow{2}{*}{\textbf{Model}} & \multirow{2}{*}{\textbf{TC}} & \multicolumn{2}{c}{\textbf{Simple}} & \multicolumn{2}{c}{\textbf{Underfitting}} & \multicolumn{2}{c}{\textbf{Overfitting}} \\ \cmidrule(lr){3-4} \cmidrule(lr){5-6} \cmidrule(lr){7-8}  
                             &                                 & \textbf{weak}    & \textbf{patch}   & \textbf{weak}       & \textbf{patch}      & \textbf{weak}       & \textbf{patch}     \\ \hline \hline
\multirow{11}{*}{SimCNN}        & airplane                                         & 78.49                             & \textbf{82.93}                     & 45.37                             & \textbf{73.61}                     & 54.29                             & \textbf{65.52}                     \\
                                & automobile                                       & 84.96                             & \textbf{95.74}                     & 62.62                             & \textbf{95.95}                     & 66.35                             & \textbf{85.00}                     \\
                                & bird                                             & 70.53                             & \textbf{85.07}                     & 33.06                             & \textbf{72.22}                     & 50.81                             & \textbf{74.65}                     \\
                                & cat                                              & 55.96                             & \textbf{64.77}                     & 25.88                             & \textbf{29.80}                     & 43.53                             & \textbf{50.67}                     \\
                                & deer                                             & 69.39                             & \textbf{80.25}                     & 42.47                             & \textbf{66.94}                     & 62.20                             & \textbf{73.53}                     \\
                                & dog                                              & 69.52                             & \textbf{84.62}                     & 35.85                             & \textbf{60.75}                     & 50.00                             & \textbf{62.16}                     \\
                                & frog                                             & 74.07                             & \textbf{77.67}                     & 34.58                             & \textbf{47.18}                     & 57.63                             & \textbf{74.74}                     \\
                                & horse                                            & 80.00                             & \textbf{89.04}                     & 63.64                             & \textbf{100.0}                     & 60.00                             & \textbf{84.62}                     \\
                                & ship                                             & 75.73                             & \textbf{81.72}                     & 100.0                             & 75.00                              & 67.29                             & \textbf{74.23}                     \\
                                & truck                                            & 74.31                             & \textbf{85.71}                     & 50.94                             & \textbf{87.84}                     & 61.26                             & \textbf{71.28}                     \\ \cline{2-8} 
                                & \textbf{Average}                                 & 73.30                             & \textbf{82.75}                     & 49.44                             & \textbf{70.93}                     & 57.34                             & \textbf{71.64}                     \\ \hline
\multirow{11}{*}{ResCNN}        & airplane                                         & 72.32                             & \textbf{84.71}                     & 40.00                             & \textbf{79.17}                     & 63.73                             & \textbf{77.03}                     \\
                                & automobile                                       & 87.25                             & \textbf{100.0}                     & 67.80                             & \textbf{100.0}                     & 56.30                             & \textbf{87.01}                     \\
                                & bird                                             & 82.93                             & \textbf{95.16}                     & 52.38                             & \textbf{82.46}                     & 52.38                             & \textbf{75.71}                     \\
                                & cat                                              & 73.75                             & \textbf{90.00}                     & 60.00                             & 56.38                              & 55.56                             & \textbf{71.70}                     \\
                                & deer                                             & 74.32                             & \textbf{92.31}                     & 36.36                             & \textbf{83.33}                     & 55.00                             & \textbf{72.73}                     \\
                                & dog                                              & 67.77                             & \textbf{77.23}                     & 36.70                             & \textbf{67.53}                     & 47.87                             & \textbf{57.69}                     \\
                                & frog                                             & 76.92                             & \textbf{87.91}                     & 44.31                             & \textbf{76.85}                     & 62.75                             & \textbf{73.81}                     \\
                                & horse                                            & 81.40                             & \textbf{92.54}                     & 39.22                             & \textbf{86.67}                     & 58.59                             & \textbf{86.15}                     \\
                                & ship                                             & 75.22                             & \textbf{84.04}                     & 50.00                             & \textbf{94.87}                     & 68.57                             & \textbf{80.00}                     \\
                                & truck                                            & 89.16                             & \textbf{95.95}                     & 28.74                             & \textbf{89.29}                     & 54.21                             & \textbf{78.95}                     \\ \cline{2-8} 
                                & \textbf{Average}                                 & 78.10                             & \textbf{89.99}                     & 45.55                             & \textbf{81.66}                     & 57.50                             & \textbf{76.08}                     \\ \hline
                                
\end{tabular}
}
\end{table}

\begin{table}
\caption{Precision of weak models and patched models on SVHN. All results in \%.}
\label{tab:patch_svhn_precision}
\vspace{-6pt}
\resizebox{\columnwidth}{!}{
\begin{tabular}{ccrrrrrr}
\hline
\multirow{2}{*}{\textbf{Model}} & \multirow{2}{*}{\textbf{TC}} & \multicolumn{2}{c}{\textbf{Simple}}                                    & \multicolumn{2}{c}{\textbf{Underfitting}}                              & \multicolumn{2}{c}{\textbf{Overfitting}}                               \\ \cmidrule(lr){3-4} \cmidrule(lr){5-6} \cmidrule(lr){7-8}
                             &                                 & \multicolumn{1}{c}{\textbf{weak}} & \multicolumn{1}{c}{\textbf{patch}} & \multicolumn{1}{c}{\textbf{weak}} & \multicolumn{1}{c}{\textbf{patch}} & \multicolumn{1}{c}{\textbf{weak}} & \multicolumn{1}{c}{\textbf{patch}} \\ \hline \hline
\multirow{6}{*}{SimCNN}         & 0                                                & 88.94                             & \textbf{95.05}                     & 25.17                             & \textbf{74.31}                     & 89.34                             & \textbf{92.24}                     \\
                                & 1                                                & 96.14                             & \textbf{96.49}                     & 92.17                             & \textbf{94.00}                     & 96.78                             & \textbf{97.39}                     \\
                                & 2                                                & 93.36                             & \textbf{96.47}                     & 95.29                             & \textbf{96.47}                     & 95.57                             & \textbf{96.42}                     \\
                                & 3                                                & 89.21                             & \textbf{95.74}                     & 89.86                             & \textbf{96.19}                     & 93.47                             & \textbf{95.24}                     \\
                                & 4                                                & 93.95                             & \textbf{97.62}                     & 87.97                             & \textbf{94.88}                     & 93.66                             & \textbf{95.34}                     \\ \cline{2-8} 
                                & \textbf{Average}                                 & 92.32                             & \textbf{96.27}                     & 78.09                             & \textbf{91.17}                     & 93.76                             & \textbf{95.33}                     \\ \hline
\multirow{6}{*}{ResCNN}         & 0                                                & 81.17                             & \textbf{93.08}                     & 78.05                             & \textbf{93.03}                     & 90.91                             & \textbf{94.24}                     \\
                                & 1                                                & 89.36                             & \textbf{95.34}                     & 95.86                             & \textbf{97.68}                     & 89.78                             & \textbf{95.44}                     \\
                                & 2                                                & 93.81                             & \textbf{97.38}                     & 95.28                             & \textbf{98.62}                     & 95.24                             & \textbf{97.23}                     \\
                                & 3                                                & 88.91                             & \textbf{95.01}                     & 80.01                             & \textbf{92.72}                     & 92.42                             & \textbf{94.93}                     \\
                                & 4                                                & 93.38                             & \textbf{98.51}                     & 57.73                             & \textbf{75.60}                     & 92.98                             & \textbf{96.18}                     \\ \cline{2-8} 
                                & \textbf{Average}                                 & 89.33                             & \textbf{95.86}                     & 81.39                             & \textbf{91.53}                     & 92.27                             & \textbf{95.60}                     \\ \hline
\end{tabular}
}
\end{table}

To answer RQ2, we conduct experiments on three common types of weak CNN models, i.e., \textit{overly simple models}, \textit{underfitting models}, and \textit{overfitting models}. 
An overly simple model has fewer parameters than a strong model.
To obtain the overly simple models, simple SimCNN and ResCNN models are used.
Specifically, a simple SimCNN contains 2 convolutional layers and 1 FC layer, while a simple ResCNN contains 4 convolutional layers, 1 FC layer, and 1 residual connection.
An underfitting model has the same number of parameters as a strong model, but is trained with a small number of epochs.
To obtain the underfitting models, the model is trained at the $\frac{n_{best}}{2}$-th epoch, which can neither well fit the training dataset nor generalize to the testing dataset. 
The accuracy of the underfitting model is low on both the training dataset and the testing dataset, indicating the occurrence of underfitting. 
An overfitting model is obtained by disableing some well-known Deep Learning ``tricks'', including dropout~\cite{dropout}, weight decay~\cite{weightdecay}, and data augmentation~\cite{dataaugmentation}. These tricks are widely used to prevent overfitting and improve the performance of a DL model. 
The overfitting model can fit the training dataset and the accuracy on the training dataset is close to 100\%; however, its accuracy on the testing dataset is much lower than that on the training dataset, indicating the occurrence of overfitting.
Except for the special design above, the same settings are applied as presented in RQ1.

We assume that 
the weak model may outperform the strong model on some non-TCs or some non-TCs may not be supported %
by the strong model.
The weak model can be improved instead of being thrown away. 
In this experiment, a weak model can recognize only the TC as well as classes that are not recognized by a strong model (see Section \ref{subsec:benchmarks}).
Therefore, each module in a strong model is used in turn as a patch to improve a weak model in recognizing the corresponding TC.
As a result, a total of 90 weak models are obtained, among which, 60 weak models for CIFAR-100 (10 classes with each class has 3 weak models for SimCNN and ResCNN, respectively) and 30 weak models for SVHN-W (5 classes with each class has 3 weak models for SimCNN and ResCNN, respectively).
Finally, given a set of overly simple, underfitting, and overfitting models, the effectiveness of using modules as patches can be validated by quantitatively and qualitatively measuring the improvement of weak models.
Specifically, the ability of a weak model to recognize a TC can be evaluated in terms of precision and recall. 
Precision is the fraction of the data belonging to TC among the data predicted to be TC.
Recall indicates how much of all data, belonging to TC that should have been found, were found. 
F1-score is used as a weighted harmonic mean to combine precision and recall.

Table \ref{tab:patch_cifar_precision} and Table \ref{tab:patch_svhn_precision} show the performance of weak models and patched models in terms of precision on CIFAR-100 and SVHN-W, respectively.
If the patched weak model performs better than the weak model, the result is in bold. 
For CIFAR-100, the results in Table \ref{tab:patch_cifar_precision} suggest that patched weak models improve precision on 100\% (20/20) of simple models, 90\% (18/20) of underfitting models, and 100\% (20/20) of overfitting models.
The average gain of precision is 18.64\%. 
The failure of the two underfitting models to improve precision could be due to their extremely low recall (i.e., 5\% and 3\%), whereas patching improves their recall (to 63\% and 53\%) rather than precision.
For SVHN-W, the results in Table \ref{tab:patch_svhn_precision} suggest that patched models improve the precision of all the simple, underfitting, and overfitting models. 
The average improvement in precision is 6.44\%.

\begin{table}[!h]
\caption{Average recall (upper half) and F1-score (bottom half) of weak models and patched models. All results in \%.}
\label{tab:recall_f1_summary}
\vspace{-6pt}
\resizebox{\columnwidth}{!}{
\begin{tabular}{ccrrrrrr}
\hline
\multirow{2}{*}{\textbf{Model}} & \multicolumn{1}{c}{\multirow{2}{*}{\textbf{Dataset}}} & \multicolumn{2}{c}{\textbf{Simple}}                                    & \multicolumn{2}{c}{\textbf{Underfitting}}                              & \multicolumn{2}{c}{\textbf{Overfitting}}                               \\ \cmidrule(lr){3-4} \cmidrule(lr){5-6} \cmidrule(lr){7-8}
                                  & \multicolumn{1}{c}{}                                & \multicolumn{1}{c}{\textbf{weak}} & \multicolumn{1}{c}{\textbf{patch}} & \multicolumn{1}{c}{\textbf{weak}} & \multicolumn{1}{c}{\textbf{patch}} & \multicolumn{1}{c}{\textbf{weak}} & \multicolumn{1}{c}{\textbf{patch}} \\ \hline \hline
\multirow{2}{*}{SimCNN} & CIFAR                                        & 74.50                    & 70.20                     & 36.90                    & \textbf{65.00}                     & 59.40                    & 57.70            \\
                        & SVHN                                         & 93.04                    & 91.31                     & 77.55                    & \textbf{85.96}                     & 92.43                    & 91.71            \\ \hline
\multirow{2}{*}{ResCNN} & CIFAR                                        & 74.30                    & 68.70                     & 39.00                    & \textbf{53.60}                     & 57.90                    & 55.80            \\
                        & SVHN                                         & 94.68                    & 90.91                     & 84.18                    & 79.14                              & 93.28                    & 91.79            \\ \hline \hline

\multirow{2}{*}{SimCNN} & CIFAR & 73.76 & \textbf{75.68} & 35.46 & \textbf{64.59} & 58.14 & \textbf{63.60} \\
                        & SVHN  & 92.67 & \textbf{93.71} & 77.71 & \textbf{88.38} & 93.08 & \textbf{93.46} \\ \hline
\multirow{2}{*}{ResCNN} & CIFAR & 75.66 & \textbf{77.13} & 34.43 & \textbf{63.05} & 57.51 & \textbf{64.09} \\
                        & SVHN  & 91.88 & \textbf{93.30} & 79.63 & \textbf{82.31} & 92.70 & \textbf{93.61} \\ \hline

\end{tabular}
}
\end{table}
The results of Recall and F1-score are summarized in Table \ref{tab:recall_f1_summary}.
As shown in the upper half of Table \ref{tab:recall_f1_summary}, on average, the gains of recall on SimCNN and ResCNN are 4.68\% and -0.57\%, respectively.
The recall values of some patched models decrease, as there is often an inverse relationship between precision and recall~\cite{inverse,fscore}. 
However, the improvement in F1-score indicates that, in general, weak models can be improved through patching. 
The bottom half of Table \ref{tab:recall_f1_summary} summarizes the results of F1-score, showing the average F1-score of weak models and patched weak models.
On average, the improvements in F1-score are 8.10\% and 6.94\% for SimCNN and ResCNN, respectively. 
Overall, the average improvements of 90 patched weak models in terms of precision, recall, and F1-score are 12.54\%, 2.14\%, and 7.52\%, respectively. 
Due to space limitation, the detailed results of the Recall and F1-score are available at the project webpage~\cite{cnnsplitter}.

\begin{table}[!h]
\caption{Average accuracy of weak models and patched models on non-TCs. All results in \%.}
\label{tab:non_tc_acc_summary}
\vspace{-6pt}
\resizebox{\columnwidth}{!}{
\begin{tabular}{ccrrrrrr}
\hline
\multirow{2}{*}{\textbf{Model}} & \multicolumn{1}{c}{\multirow{2}{*}{\textbf{Dataset}}} & \multicolumn{2}{c}{\textbf{Simple}}                                    & \multicolumn{2}{c}{\textbf{Underfitting}}                              & \multicolumn{2}{c}{\textbf{Overfitting}}                               \\ \cmidrule(lr){3-4} \cmidrule(lr){5-6} \cmidrule(lr){7-8}
                                  & \multicolumn{1}{c}{}                                & \multicolumn{1}{c}{\textbf{weak}} & \multicolumn{1}{c}{\textbf{patch}} & \multicolumn{1}{c}{\textbf{weak}} & \multicolumn{1}{c}{\textbf{patch}} & \multicolumn{1}{c}{\textbf{weak}} & \multicolumn{1}{c}{\textbf{patch}} \\ \hline \hline
\multirow{2}{*}{SimCNN} & CIFAR                                        & 77.45                    & \textbf{78.23}            & 42.72                    & \textbf{43.31}            & 58.59                    & \textbf{59.47}            \\
                        & SVHN                                         & 88.88                    & \textbf{89.94}            & 73.97                    & \textbf{75.13}            & 91.14                    & \textbf{91.52}            \\ \hline
\multirow{2}{*}{ResCNN} & CIFAR                                        & 81.16                    & \textbf{81.96}            & 42.80                    & \textbf{44.53}            & 60.25                    & \textbf{61.28}            \\
                        & SVHN                                         & 88.06                    & \textbf{90.14}            & 74.99                    & \textbf{78.82}            & 90.36                    & \textbf{91.53}            \\ \hline

\end{tabular}
}
\end{table}

Besides the improvement in recognizing TC, 
another concern is whether the patch affects the ability to recognize other classes (i.e., non-TCs).
To evaluate the patch's effects on non-TCs, the samples belonging to TC are removed, and weak models and patched models are evaluated on the samples belonging to non-TCs.
Finally, the effect of the patch on non-TCs is validated by comparing the accuracy of weak models to patched models. 
The experimental results~\cite{cnnsplitter}  are summarized in Table \ref{tab:non_tc_acc_summary}.
Overall, 94\% (85/90) of patched models outperform the weak models, and the average accuracy improvement of 90 patched models is 1.18\%. 
The reason for performance improvement is that some samples that belong to non-TCs but were misclassified as TC are correctly classified as non-TCs after patching.
The results indicate that the patching does not impair but rather improves the ability to recognize non-TCs. 

Moreover, for the costs of prediction, patching does not incur extra time costs, as the patch and the weak CNN model are executed in parallel. 
The GPU memory consumption of the weak and patched CNN models is about 1.7GB and 2.6GB, respectively. 
Although the patch incurs extra GPU memory consumption, the overhead of reusing a module as a patch is much lower than that of reusing an entire model. 
\begin{tcolorbox}
\vspace{-6pt}
The average improvements on TC in precision, recall, and F1-score are 12.54\%, 2.14\%, and 7.52\%, respectively.
Also, the average accuracy improvement on non-TCs is 1.18\%. 
Experimental results demonstrate that patching can improve weak models not only on TC but also on non-TCs.
\vspace{-15pt}
\end{tcolorbox}

\ 

\noindent\textbf{\textit{RQ3: How effective are the three heuristic methods in \projectName?}}

In RQ3, to evaluate the effectiveness of the proposed importance-based grouping and sensitivity-based initialization, 
\projectName executes with different grouping (No, Random, Importance) and initialization (Random, Sensitivity) methods.
Each trained CNN model is modularized with these configurations. 
To validate the effectiveness of pruning-based evaluation, \projectName executes with and without pruning. 
To measure the effectiveness of these heuristics, we compare \textit{Acc} and which generation obtains the best modules under different configurations.

\begin{table}[]
\caption{The results of modularization in different configurations. }%
\label{tab:RQ2}
\vspace{-6pt}
\begin{center}
{
\resizebox{\columnwidth}{!}{
\begin{tabular}{ccccr}
\toprule
\multirow{2}{*}{\textbf{Model Name}} & \multicolumn{2}{c}{\textbf{Modularization Methods}} & \multicolumn{2}{c}{\textbf{Composed Model}} \\ \cmidrule(lr){2-3} \cmidrule(lr){4-5}
                                & \textbf{Grouping}       & \textbf{Initialization}     & \textbf{Generation}      & \textbf{Acc}        \\ \hline \hline
\multirow{4}{*}{SimCNN-CIFAR}   & No                      & Sensitivity                   & 194                      & 0.2754              \\
                                & Random                  & Sensitivity                   & 190                      & 0.3650              \\
                                & Importance               & Random                      & 192                      & 0.3702              \\ \cline{2-5} 
                                & \textbf{Importance}      & \textbf{Sensitivity}          & \textbf{123}             & \textbf{0.8607}     \\ \hline
\multirow{4}{*}{SimCNN-SVHN}    & No                      & Sensitivity                   & 200                      & 0.2430              \\
                                & Random                  & Sensitivity                   & 200                      & 0.2512              \\
                                & Importance               & Random                      & 188                      & 0.9204              \\ \cline{2-5} 
                                & \textbf{Importance}      & \textbf{Sensitivity}          & \textbf{79}              & \textbf{0.9385}     \\ \hline
\multirow{4}{*}{ResCNN-CIFAR}   & No                      & Sensitivity                   & 83                       & 0.7271              \\
                                & Random                  & Sensitivity                   & 193                      & 0.8420              \\
                                & Importance               & Random                      & 197                      & 0.8432              \\ \cline{2-5} 
                                & \textbf{Importance}      & \textbf{Sensitivity}          & \textbf{185}             & \textbf{0.8564}     \\ \hline
\multirow{4}{*}{ResCNN-SVHN}    & No                      & Sensitivity                   & 140                      & 0.9027              \\
                                & Random                  & Sensitivity                   & 162                      & 0.9249              \\
                                & Importance              & Random                      & 179                      & 0.9332              \\ \cline{2-5} 
                                & \textbf{Importance}      & \textbf{Sensitivity}          & \textbf{107}             & \textbf{0.9352}     \\ \hline
\end{tabular}
}
}
\end{center}
\vspace{-6pt}
\end{table}

\textbf{Importance-based grouping.}
Table \ref{tab:RQ2} shows the evaluation results under different configurations.
For SimCNN-CIFAR and SimCNN-SVHN, modularization fails with no grouping and random grouping due to the significant loss of accuracy. 
For ResCNN-CIFAR, the modularization also fails with no grouping.
Although the modularization with random grouping is successful on ResCNN-CIFAR, the results in terms of \textit{Acc} and generation are much worse than importance-based grouping.
Similarly, no grouping and random grouping work on ResCNN-SVHN; however, both they lose more accuracy and require a larger number of generations when compared to the importance-based grouping.
We believe that our importance-based grouping can significantly improve search efficiency by reducing the size of the search space.

\textbf{Sensitivity-based initialization.}
As shown in Table~\ref{tab:RQ2}, the modularization with the random initialization fails on SimCNN-CIFAR due to the significant loss of accuracy.
While on SimCNN-SVHN, ResCNN-CIFAR, and ResCNN-SVHN, the random initialization is successful; however, compared to the sensitivity-based initialization, the accuracy decreases slightly, and the number of generations increases a lot. 
Sensitivity-based initialization method can significantly improve the efficiency of the search by providing a good starting point.

\textbf{Pruning-based evaluation.}  
The pruning strategy can significantly reduce the time complexity of module evaluation. 
In the absence of the pruning strategy, there are a total of $100^{10}$ CMs to be evaluated in each generation. 
The number of CMs is so large that each generation takes several years to evaluate, resulting in a timeout and modularization failure. 
With the pruning strategy, a 10-class classification task is decomposed into five 2-class classification tasks. Four of five 2-class classification tasks are composed into two 4-class classification tasks that are then composed into an 8-class classification task. 
The 8-class classification task and the remaining 2-class classification task are composed into the final 10-class classification task. 
Consequently, the number of CMs is $5 \times 100^2 + 2 \times (N_{top})^2 + (N_{top})^2 + (N_{top})^2$, which is much less than $100^{10}$. 
For instance, when $N_{top}=100$, the number of CMs is $9 \times 100^2$, meaning that the pruning strategy reduces time complexity from $O(n^{10})$ to $O(n^2)$.
This strategy enables the computation of accuracy to be completed with acceptable overhead. 
In the experiments, the time overhead per generation with pruning for SimCNN-CIFAR, SimCNN-SVHN, ResCNN-CIFAR, and ResCNN-SVHN is 83s, 95s, 80s, and 93s, respectively.
\begin{tcolorbox}
\vspace{-6pt}
In summary, the modularization results of the four models demonstrate that importance-based grouping, sensitivity-based initialization, and pruning-based evaluation can significantly improve the efficiency of the search.
\vspace{-6pt}
\end{tcolorbox}

\section{Discussion}
\subsection{Compressed modularization \textit{vs.} uncompressed modularization}
\label{subsec:struectued}
\begin{figure}
    \centering
    \includegraphics[width=8.0cm]{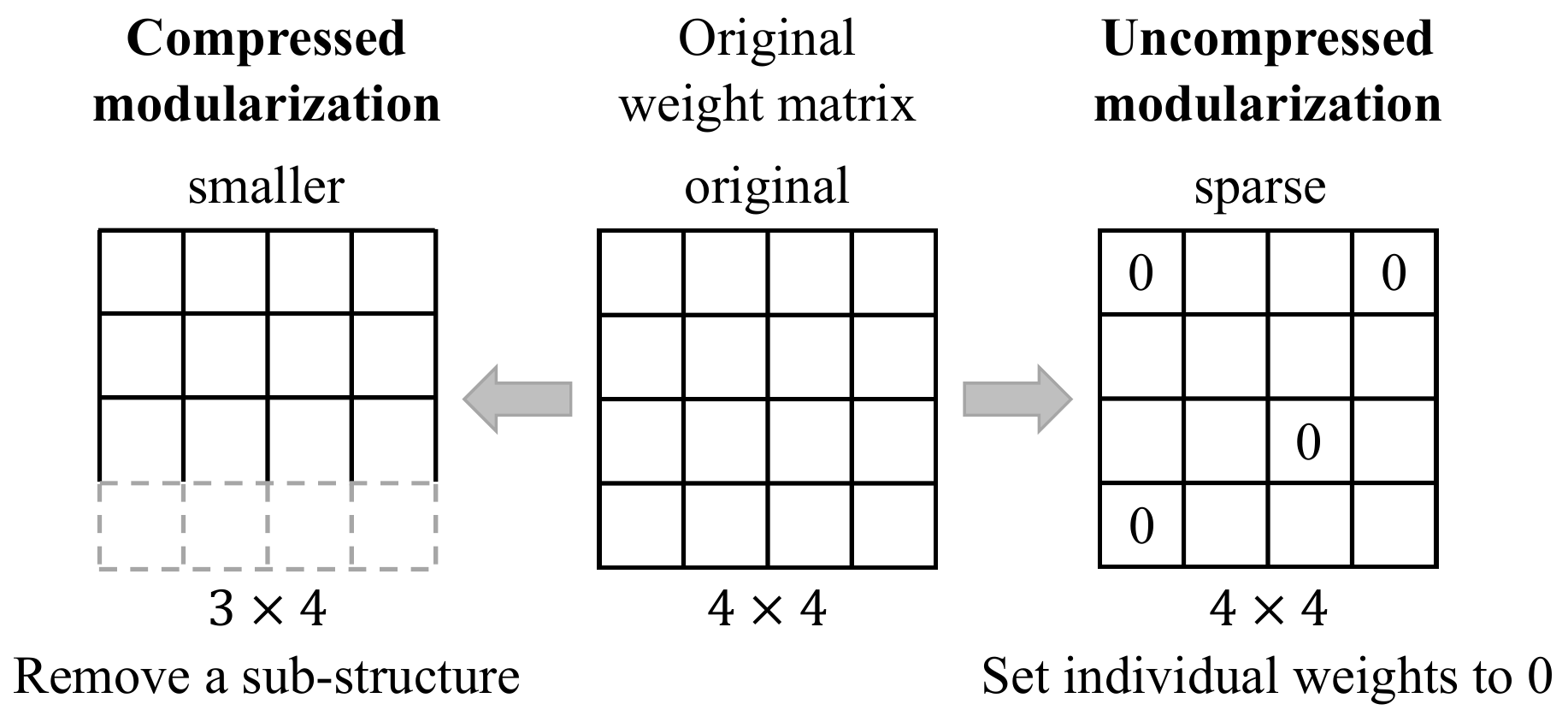}
    \caption{Compressed modularization \textit{vs.} uncompressed modularization.}
    \label{fig:illustration_structured}
    \vspace{-12pt}
\end{figure}

\projectName is a compressed modularization approach, while the existing approaches~\cite{nnmodularity2022icse,fse2020modularity} are based on uncompressed modularization. 
As shown in Figure \ref{fig:illustration_structured}, compressed modularization can remove a sub-structure from a module's weight matrix, resulting in a smaller weight matrix. In a CNN model, the convolution kernel is a sub-structure. As demonstrated in Section \ref{subsec:result} (RQ1), by removing convolution kernels, the number of weights in a module can be reduced, and hence the computation cost of a module is less than that of the original model. 

In contrast, uncompressed modularization~\cite{nnmodularity2022icse,fse2020modularity} sets individual neurons or weights to 0, thereby achieving the effect of removing the neurons or weights from CNN models to produce modules. 
As shown in Figure \ref{fig:illustration_structured}, a module produced by uncompressed modularization has the same number of weights as the original model.
As discussed in Section \ref{subsec:result}, the computation cost of a module produced by~\cite{fse2020modularity} is the same as that of the original model.

The difference between compressed and uncompressed modularization discussed above has large impact on module reuse overhead, which is an important concern for the  modularization of DNN.
The module reuse overhead is the memory and computation cost for reusing modules, which can be measured by the number of weights and FLOPs. 
As shown in Table \ref{tab:RQ1} and Table \ref{tab:flops}, the modules produced by \projectName have fewer kernels and FLOPs than the original model, indicating that
compressed modularization incurs less module reuse overhead than uncompressed modularization.

\subsection{Threats to Validity}
\label{sec:threats}
\textbf{External validity:} Threats to external validity relate to the generalizability of our results. 
There are some CNN models that have different structures from that of SimCNN and ResCNN. 
The results might not be generalizable to these CNN models. 
Moreover, the results are not validated on other datasets for different tasks. 
However, SimCNN and ResCNN are representative CNN models and their structures are widely used in various tasks. 
Many different CNN models can be seen as variants of SimCNN and ResCNN. 
In addition, CIFAR-10, CIFAR-100, and SVHN are representative datasets and are widely used for evaluation in related researches~\cite{nnmodularity2022icse,feng2020deepgini}. 
Moreover, an assumption behind our work is that only a part of classes of weak model are overlapped with the classes of a strong model, or the strong model only outperforms the weak model on part of the classes.
For the scenarios where the classes recognized by the strong model is a superset of the classes recognized by the weak model and the strong model performs better on all the classes, developers could directly use the strong model instead of patching the weak model.
In future work, more experiments will be conducted on a variety of models and datasets to alleviate this threat.

\textbf{Internal validity:} An internal threat comes from the choice of datasets for evaluation. To mitigate this threat, in this study we use the CIFAR-10, CIFAR-100, and SVHN datasets 
from Pytorch, which are well organized and widely used.

\textbf{Construct validity:} In this study, a threat relates to the suitability of our evaluation metrics. We use accuracy and Jaccard distance-based difference as the evaluation metrics. These metrics have also been used in other related work~\cite{fse2020modularity,nnmodularity2022icse}.

\section{Related Work}
\label{sec:related_work}
\textit{Uncompressed modularization of DNNs:}
The closest work to ours is uncompressed modularization~\cite{fse2020modularity,nnmodularity2022icse,nnmodularity2021arxiv}, which decompose a DNN model into modules by removing weights or neurons.
Pan et al.~\cite{fse2020modularity} first proposed determining whether neurons and weights are relevant for recognizing a specific class according to whether the neurons are activated and generating a module by removing the weights and neurons that are irrelevant for recognizing the specific class. 
However, \cite{fse2020modularity} is unsuitable for CNN models due to the weight sharing in CNNs~\cite{cnn2018overview}. 
To modularize CNN models, Pan et al.~\cite{nnmodularity2022icse} used a map to record at which positions a convolution kernel produces irrelevant neurons (i.e., inactive neurons) in the modularization phase and set the values of irrelevant neurons to zero in the prediction phase. 
Kingetsu et al.~\cite{nnmodularity2021arxiv} applied the edge-popup algorithm~\cite{edgepopup} to learn a supermask for each module that records which weights in the trained model are retained by a module. 
These uncompressed modularization methods produce modules with the same size of weight matrices as the original model, which could %
introduce large overhead in module reuse.
In contrast, our work is compressed modularization, 
which generates modules with smaller weight matrices,
leading to less overhead in module reuse. %

\textit{Reusing neural networks:} 
Our work is related to the work on reusing neural networks, such as transfer learning~\cite{transfer_learning, devlin2018bert} and conditional computation~\cite{yang2019condconv,kirsch2018modular}. 
Transfer learning techniques develop a new model by reusing the entire or a part of a model trained on the other dataset and then fine-tuning the reused model on the new dataset. For instance, BERT~\cite{devlin2018bert} is widely-used to develop new models for various downstream tasks by changing the heads (i.e., the output layers) and fine-tuning on the new dataset. 
Conditional computation aims to increase model capacity without a proportional increase in computation cost. %
For instance, Kirsch et al.~\cite{kirsch2018modular} construct a dynamic network structure consisting of modules and controllers, which activates the different portion of the entire network for the different input.  %
The techniques mentioned above attempt to reuse; however, 
they either need to retrain (fine-tuning) the reused model ~\cite{devlin2018bert} or cannot be used to develop a new model~\cite{yang2019condconv,kirsch2018modular}. 
In contrast, this work modularizes a CNN model so that a module can be reused as a patch to improve a weak CNN model without retraining. 

\textit{DNN Debugging:} 
Existing DNN debugging techniques improve DNNs mainly by providing more training data~\cite{ma2018mode}. 
One of the mainstream DNN debugging techniques is the generation technique~\cite{xie2019deephunter, tian2018deeptest, zhang2018deeproad, chen2016infogan}, which generates new training samples that are similar to the provided input data samples. 
For instance, DeepHunter~\cite{xie2019deephunter} and DeepTest~\cite{tian2018deeptest} generate new images by mutating an original image with metamorphic mutations such as pixel value transformation and affine transformation. 
DeepRoad~\cite{zhang2018deeproad} and infoGAN~\cite{chen2016infogan} are based on the generative adversarial network, which train a generator and a discriminator, and use the generator to generating new images.  
Another popular technique is the prioritization technique~\cite{wang2021prioritizing, feng2020deepgini}, which can find the possibly-misclassified data from massive unlabeled data. 
The possibly-misclassified data, rather than total data, are manually labeled first and added into the training dataset to improve a DNN.
Unlike the existing DNN debugging techniques focusing on retraining models with more training data, this work focuses on patching models without retraining.

\textit{Neural Architecture Search:}
Neural architecture search (NAS) techniques~\cite{genetic2017,genetic2019} construct the optimal neural network structure by searching combinations of network layers, layer connections, activation methods, and so on. %
\projectName searches modules from a trained CNN model.
Apart from their differences in objectives, there are some other differences between \projectName and NAS. 
For instance, genetic CNN~\cite{genetic2017} encodes CNN model architectures into bit vectors and applies a genetic algorithm to search. Each bit of bit vectors represents whether or not a connection between two convolutional layers is required. %
While in \projectName, each bit of bit vectors represents whether a kernel group is retained; thus, the approach of genetic CNN cannot be directly applied to modularization. Moreover, \projectName includes three heuristic methods (see Section \ref{sec:approach}) to improve the efficiency of search, which are also different from genetic CNN.

\section{Conclusion}
\label{sec:conclusion}
In this work, we explore how a CNN model can be decomposed into a set of smaller and reusable modules so that a module from a strong CNN model can be used as a patch to improve a weak CNN model without retraining. 
We propose a compressed modularization
approach, \projectName, to address the modularity issue of CNN models with a search-based method. 
The obtained modules can be used as patches to improve the performance of weak CNN models. 
We have evaluated \projectName with two representative CNN models on three widely-used datasets. 
The experimental results confirm the effectiveness of our approach.

Our source code  and experimental data are available at \textbf{\url{https://github.com/qibinhang/CNNSplitter}}.

\section*{Acknowledgement}
This work was supported partly by National Natural Science Foundation of China under Grant Nos.(62141209, 61932007, 61972013) and by Australian Research Council Discovery Projects (DP200102940, DP220103044) and sponsored by Huawei Innovation Research Plan.

\balance
\bibliographystyle{ACM-Reference-Format}
\bibliography{reference}

\end{document}